\documentclass[conference]{IEEEtran}
\newcommand{\eename}{E$^2$CM}
\usepackage{cite}
\usepackage{amsmath,amssymb,amsfonts}
\usepackage{algorithmic}
\usepackage{algorithm}
\usepackage{graphicx}
\usepackage{textcomp}
\usepackage{xcolor}
\usepackage{multirow}

\def\BibTeX{{\rm B\kern-.05em{\sc i\kern-.025em b}\kern-.08em
    T\kern-.1667em\lower.7ex\hbox{E}\kern-.125emX}}
\begin{document}

\title{ \eename{}: Early Exit via Class Means for Efficient Supervised and Unsupervised Learning}

%\author{\IEEEauthorblockN{Anonymous Authors}}

\IEEEoverridecommandlockouts
\IEEEaftertitletext{\vspace{-10pt}}

\author{\IEEEauthorblockN{Alperen Görmez}
\IEEEauthorblockA{\textit{University of Illinois at Chicago}\\
agorme2@uic.edu}
\and
\IEEEauthorblockN{Venkat R. Dasari}
\IEEEauthorblockA{\textit{CCDC US Army Research Laboratory} \\
venkateswara.r.dasari.civ@army.mil}
\and
\IEEEauthorblockN{Erdem Koyuncu}
\IEEEauthorblockA{\textit{University of Illinois at Chicago}\\
ekoyuncu@uic.edu }
}

\maketitle

\begin{abstract}
State-of-the-art neural networks with early exit mechanisms often need considerable amount of training and fine tuning to achieve good performance with low computational cost. We propose a novel early exit technique, \emph{Early Exit Class Means (\eename{})}, based on class means of samples. Unlike most existing schemes, \eename{} does not require gradient-based training of internal classifiers and it does not modify the base network by any means. This makes it particularly useful for neural network training in low-power devices, as in wireless edge networks. We evaluate the performance and overheads of \eename{} over various base neural networks such as MobileNetV3, EfficientNet, ResNet, and datasets such as CIFAR-100, ImageNet, and KMNIST. Our results show that, given a fixed training time budget, \eename{} achieves higher accuracy as compared to existing early exit mechanisms. Moreover, if there are no limitations on the training time budget, \eename{} can be combined with an existing early exit scheme to boost the latter's performance, achieving a better trade-off between computational cost and network accuracy. We also show that \eename{} can be used to decrease the computational cost in unsupervised learning tasks.
\end{abstract}
% *CRITICAL: Do Not Use Symbols, Special Characters, Footnotes, or Math in Paper Title or Abstract.

\begin{IEEEkeywords}
Neural networks, early exit, class means.
\end{IEEEkeywords}

\section{Introduction}
Modern deep learning models require a vast amount of computational resources to effectively perform various tasks such as object detection \cite{szegedy2015going}, image classification \cite{he2016deep}, machine translation \cite{vaswani2017attention} and text generation \cite{brown2020languagegpt3}. Deploying deep learning models to the edge, such as to mobile phones or the Internet of Things (IoT), thus becomes particularly challenging due to device computation and energy limitations \cite{respipe, li2021model}. Moreover, the law of diminishing returns applies to the computation-performance trade-off \cite{bolukbasi}: The increase in the performance of a deep learning model is often marginal as compared to the increase in the amount of computation.

% , and the possible simplicity of an input is not exploited

One of the primary reasons behind traditional deep learning models' high computation demand is their tunnel-like design. In fact, traditional models apply the same sequence of operations to any given input. However, in many real world datasets, certain inputs may consist of much simpler features as compared to other inputs \cite{bolukbasi}. In such a scenario, it becomes desirable to design more efficient architectures that can exploit the heterogeneous complexity of dataset members. This can be achieved by introducing additional exit points to the models \cite{panda2016cdl, branchynet, shallowdeep, biasielli2020neural, zhou2020bert, xin2020deebert}. These exit points prevent simple inputs to traverse the entire network, reducing the computational cost of inference.

\begin{figure}
\centerline{\includegraphics[width=\linewidth]{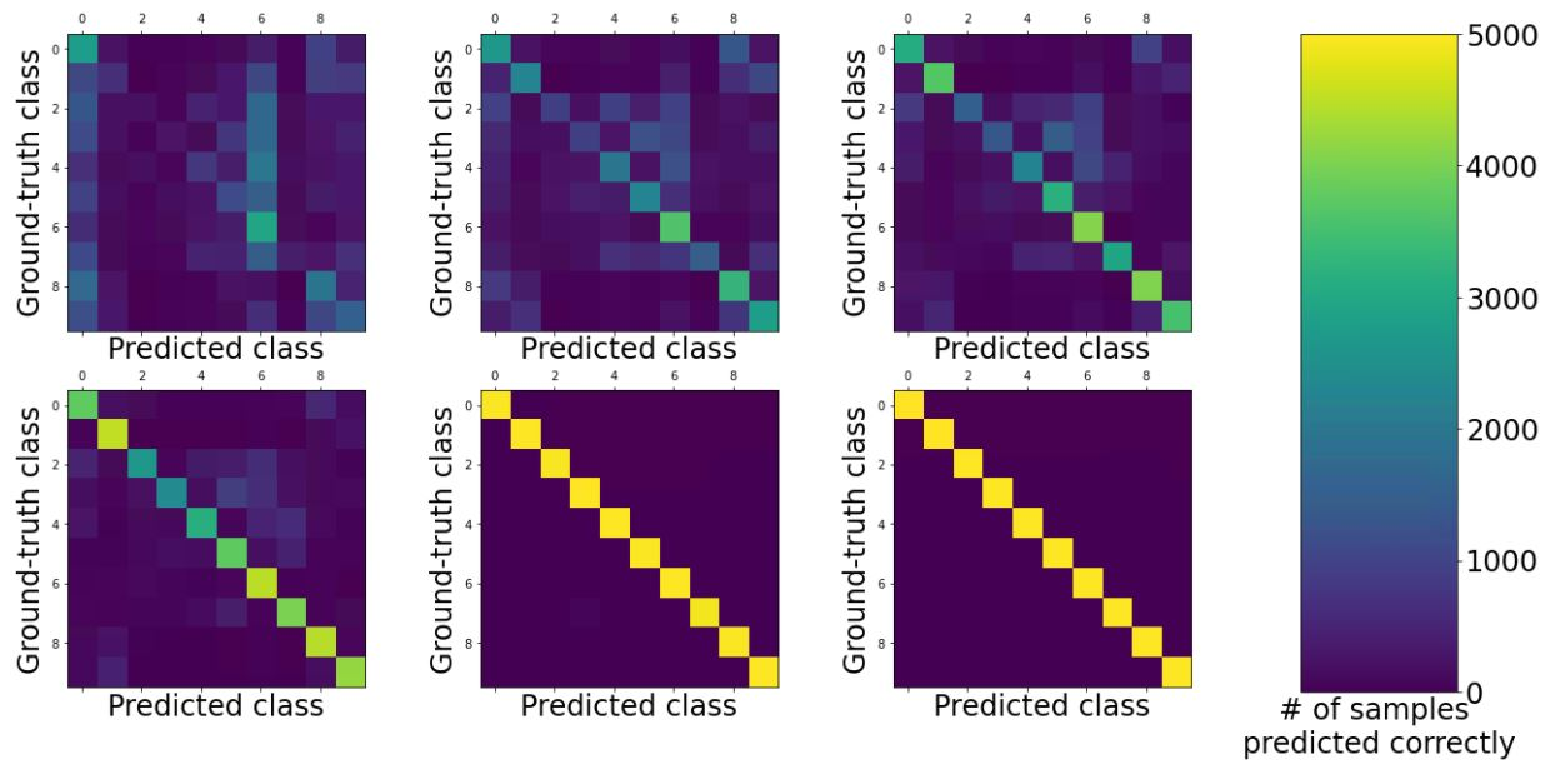}}\vspace{-10pt}
\caption{Confusion matrix of the classifications done according to the nearest class mean on CIFAR-10 training set. From left to right, top to bottom; the results belong to the first convolutional layer, 1$^{st}$, 3$^{rd}$, 10$^{th}$, 15$^{th}$ and 30$^{th}$ residual block of ResNet-152.}\vspace{-10pt}
\label{confusionmatrix}
\end{figure}

Despite reduced inference time, existing early exit neural network architectures require additional training and fine-tuning for the early exit points, which increases the training time \cite{branchynet, bolukbasi, shallowdeep}. This side-effect is undesirable for scenarios in which the training has to be done in a low-power device. An ideal solution is a plug-and-play approach that does not require gradient-based training and performs well. In this work, we propose such an early-exit mechanism, Early Exit Class Means (\eename{}), based on the \emph{class means} of input samples for the image classification task. By averaging the layer outputs for each class at every layer of the model, class means are obtained. During inference, output of a layer is compared with the corresponding class means using Euclidean distance as the metric. If the output of the layer is close enough to a class mean, the execution is stopped and the sample exits the network. In fact, as seen in Fig.~\ref{confusionmatrix}, some samples can be classified easily at early stages of the network by just considering a ``nearest class mean'' decision rule, suggesting the potential effectiveness of our method for reducing computational cost.

A practical use case for \eename{} is when a large, expensive-to-train model is broadcast to edge devices with limited and heterogeneous computation capabilities. In such a scenario, different devices may train the \eename{} model at different FLOP operating points depending on their computation capabilities or power limitations.

One other practical scenario for \eename{} is transfer learning and fine tuning, where the training has to be done on low-power edge device with a local dataset that is different from the dataset the base network was trained on. In order to keep the battery usage minimum on the edge device, the training time may be limited. The advantage of \eename{} is that it does not need any gradient-based training like Shallow-Deep Networks \cite{shallowdeep}, therefore it is more suitable for transfer learning and fine tuning on low-power devices.

To the best of our knowledge, \eename{} is the first early exit mechanism that does not require gradient-based training and does not modify the original network by any means. Moreover, \eename{} does not have a hyper-parameter for early exit locations unlike existing schemes. These features make \eename{} simpler to use and easier to deploy on low-power devices. While using class means as the only early exit mechanism requires just a single feed forward pass, existing early exit methods reaches the same performance after training for multiple epochs (i.e. multiple forward and backward passes), which suggests our method is more agile and powerful yet simpler. Moreover, combining \eename{} with the existing mechanisms that require gradient-based training achieves a better trade-off in terms of computation cost and network accuracy. In addition to its benefits in the supervised learning setting, \eename{} is also the first technique that shows the feasibility of early exits in the unsupervised learning setting.

We show the effectiveness of \eename{} on CIFAR-10 \cite{cifar}, CIFAR-100 \cite{cifar}, ImageNet \cite{imagenet}, Tiny ImageNet \cite{tinyimagenet}, KMNIST \cite{kmnist}, Fashion-MNIST \cite{fmnist} and MNIST \cite{mnist} datasets using ResNet-18 \cite{he2016deep}, ResNet-152 \cite{he2016deep}, WideResNet-101 \cite{zagoruyko2016wide}, MobileNetV3Large \cite{mobilenetv3} and EfficientNet-B0 \cite{efficientnet} models. Using class means as the sole decision mechanism results in 50\% better accuracy or 50\% faster inference time compared to the existing early exit techniques in the literature.  Also, when combined with the state-of-the-art, we increase the accuracy by 6\% without doing further computation; or decrease the inference time by 33\% with a negligible loss in accuracy. We also show that it is possible to decrease the computational cost while doing clustering with autoencoders on MNIST and Fashion-MNIST \cite{mnist, fmnist}. In particular, \eename{} saves on computation by 60\%, while the loss in unsupervised clustering accuracy \cite{dec} is marginal. We also evaluate the computational and memory overheads of \eename{}, and show that they are practically low even for large datasets and/or models.

% \eename{} requires storing the class means of samples, which translates to extra memory overhead. We show that, even for datasets with a large number of classes, such as ImageNet, the extra memory overhead is at most around 10\% of the base neural network model size, which is reasonably small.

% In addition, \eename{} has a reasonably small memory overhead.

\section{Related Work}
\textbf{Conditional computation.} Our work is related to the area of conditional computation \cite{bengio2013estimating, bengio2015conditional, veit2018convolutional}, where several small networks are trained to control the computation flow of one deep neural network. For this purpose, adding gates between the blocks of residual networks have been proposed \cite{veit2018convolutional, skipnet}. During inference, these gates allow the input to skip unnecessary blocks, thus saving computation time. However, the gates have to be trained from scratch jointly with the base network. Also, the locations of the gates have to be determined explicitly, which result in increased number of hyper-parameters. Unlike these methods, \eename{} does not modify the base network and does not require gradient-based training.

\textbf{Early exit networks.} One of the earliest works that explicitly propose the idea of early exiting is \cite{panda2016cdl}, where the authors consider adding a cascade of linear layers after convolutional layers as control blocks. Rather than just linear layers, adding \emph{branches} consisting of convolutional layers to the original model has also been studied \cite{branchynet}. A significant drawback of this method is that the branches increase the computational cost due to convolutional layers. In a more recent study, \emph{internal classifiers (ICs)} consisting of a feature reduction layer and a single linear layer are added after certain layers in the network \cite{shallowdeep, xin2020deebert}. The methods presented in these studies modify the original network by adding linear or convolutional layers. Moreover, they require gradient updates to train those layers. Also, an implicit hyper-parameter is the locations of the early exit points. \eename{} is better suited for low-power applications compared to existing studies since we do not modify the original model, do not require gradient based training and additional hyper-parameters. Another novelty of \eename{} is the ability to extend to unsupervised learning tasks. Existing early exit methods focus only on supervised learning.

In addition to layer level early exits, a network level early exit mechanism has been introduced in \cite{bolukbasi}. Both the layer level exit and the network level exit require decision functions to be inserted between the layers and networks. This type of architecture freezes the weights of the original model, and then trains the decision functions one by one using weighted binary classification. The drawback of this approach is such an alternative optimization, which may consume a lot of time and energy when there are many decision functions to optimize.

\textbf{Multi-resolution networks.} One other idea to reduce the inference time computational cost of neural networks is the usage of multi-resolution features to facilitate early exiting \cite{multiscaledensenet, resadaptivenets, dynrouting}. While this idea works well, these methods focus only on supervised learning tasks and they design new architectures for that purpose. Moreover, the locations of early exits are restricted to the last few blocks of the sub-network \cite{resadaptivenets}. Rather than designing new architectures, \eename{} is a plug-and-play method: We focus on taking off-the-shelf networks such as MobileNetV3Large \cite{mobilenetv3} and reducing the inference time without modifying and retraining the model for both supervised and unsupervised learning tasks.

\textbf{Few-shot learning.} Intermediate layer outputs have been used for classification in few-shot and one-shot learning settings in the past \cite{koch2015siamese, vinyals2016matching, snell2017prototypical, sung2018learning}. These studies are closely related to the area of \emph{metric learning}. The closest work to \eename{} is \emph{prototypical networks}, in which \emph{prototypes} for each class are computed \cite{ snell2017prototypical, protadapt}. However, none of those approaches aim to reduce the computation cost, rather they either try to remedy the problem of classifying unseen classes or explore unsupervised domain adaptation \cite{fewshot, protadapt}.

\textbf{Neural collapse.} \eename{} is also related to the phenomenon of neural collapse \cite{neuralcollapse}. It is known that as the inputs go deeper in a neural network, the classes are separated better from each other as a result of multiple nonlinearities, and the samples begin to concentrate \cite{separability, concentration}. \eename{} exploits this phenomenon with the main idea of stopping the execution as soon as the sample is close enough to a concentration point i.e., a class mean.

% We primarily study the problem of image classification, but the idea can be adapted to other modalities such as text and audio classification. 

\section{Early Exit Class Means}
In this section, we introduce \eename{}  in the context of image classification. The scheme extends to different classification tasks in the same manner. Let ${(x_0^{(i)},y^{(i)})} \in {D}$ be an image-label pair from the dataset ${D}$ consisting of ${N}$ samples and ${K}$ distinct classes, where ${y^{(i)} \in \{1,2,\ldots,K\}}$ and ${i \in \{1,2,\ldots,N\}}$. We denote the network ${F}$ with ${M}$ layers as a sequence ${l_1},{l_2},\ldots,{l_M}$. Let ${\hat y^{(i)}}$ denote the prediction of the network, ${x_j^{(i)}}$ denote the output of layer ${j}$, and ${\hat y_j^{(i)}}$ denote the prediction in case the input exits the network after layer ${j}$, for ${j=1,2,\ldots,M}$. The input-output relationships of the network ${F}$ can be expressed as
\begin{equation}
    {x_{j}^{(i)}=l_j(x_{j-1}^{(i)})},\, {j=1,2,\ldots,M}.
    \label{equation1}
\end{equation}

\subsection{Class Means}
The input to \eename{} is the network ${F}$ trained on ${D}$. The network ${F}$ is not modified by any means. Therefore, we can obtain the class means for each class at each layer easily by just a forward pass. This is especially useful when the training time budget is fixed. Let ${S_k}$ denote the set of samples whose ground-truth label is ${k}$, and ${c^{k}_j}$ denote the mean of the output of layer ${j}$ for class ${k}$. In other words, let
\begin{equation}
    \textstyle {c^{k}_j = \frac{1}{|S_k|} \sum_{n\in S_k}x_{j}^{(n)}}.
    %\quad \text{such that} \quad {y^{(n)} = k}.
    \label{equation2}
\end{equation}
The Euclidean distance between a layer output ${x_j^{(i)}}$ and ${K}$ class means ${c^{k}_j}$ is then computed via
\begin{equation}
    {d^{k^{(i)}}_j = ||x_j^{(i)} - c^{k}_j||_2} ,\,k \in \{1,2,\ldots,K\}.
    \label{equation3}
\end{equation}

After calculating ${d^{k^{(i)}}_j}$ at each layer for every sample in the dataset, we normalize the distances for each class as
\begin{equation}
    {d^{k^{(i)}}_j := \frac{d^{k^{(i)}}_j}{\frac{1}{N}\sum_{i=1}^{N}d^{k^{(i)}}_j}
    },\, k \in \{1,2,\ldots,K\}.
    \label{equation4}
\end{equation}

Finally, the normalized distances are converted to probabilities of input belonging to a class in order to perform inference. This is done using the softmax function as
\begin{equation}
    {P(\hat y_j^{(i)} = k) = \mathrm{softmax}(-d^{k^{(i)}}_j)}.
    \label{equation5}
\end{equation}

During inference, the decision of exiting after ${l_j}$ or moving forward to ${l_{j+1}}$ is made according to a threshold value ${T_j}$. If the largest softmax probability is greater than the specified threshold ${T_j}$, execution is stopped and the class with the largest softmax probability is predicted. In other words, if
\begin{equation}\max(\mathrm{softmax}(-d^{k^{(i)}}_j)) > T_j,\end{equation} 
then the network predicts 
\begin{equation}
\hat y_j^{(i)} = \arg\max_k(\mathrm{softmax}(-d^{k^{(i)}}_j)). 
\label{equation6}
\end{equation}

\begin{algorithm}[]
   \caption{Early Exit Class Means (\eename{})}
   \label{alg:classmeans}
\begin{algorithmic}
   \STATE {\bfseries Input:} Trained network layers $l_j$, dataset $D$, thresholds $T_j$
   \IF{training}
   \FOR{$j=1$ {\bfseries to} $M$}
   \STATE $x_{j}^{(i)}=l_j(x_{j-1}^{(i)})$
   \STATE Calculate class means $c^{k}_j,\, k \in \{1,2,\ldots,K\}$
   \ENDFOR
   \ENDIF
   
   \IF{inference}
   \FOR{$j=1$ {\bfseries to} $M$}
   \STATE $x_{j}^{(i)}=l_j(x_{j-1}^{(i)})$
   \STATE Compute $d^{k^{(i)}}_j = ||x_j^{(i)} - c^{k}_j||_2$
   \STATE Normalize $d^{k^{(i)}}_j$ as in (\ref{equation4}).
   \IF{ $\max(\mathrm{softmax}(-d^{k^{(i)}}_j)) > T_j$}
   \STATE Early exit with $ \arg\max_k(\mathrm{softmax}(-d^{k^{(i)}}_j))$.
   \ENDIF
   \ENDFOR
   \ENDIF
\end{algorithmic}
\end{algorithm}

Otherwise, the input moves forward to the next layer. In the worst case, execution ends at the last layer of the network. The full procedure of our method is shown in Algorithm~\ref{alg:classmeans}.

\eename{} performs differently according to different set of threshold values. We use binary search to reach target number of FLOPs on training set. If the thresholds result in a higher number of FLOPs than the target number of FLOPs, they are decreased to encourage early exits. Else, they are increased to encourage moving deeper in the layers. Later, same threshold values are used on the test set during the inference phase for that target number of FLOPs.

\subsection{Combination of Class Means with Existing Schemes}
\eename{} can boost the performance of existing early exit schemes. Existing methods use only ${x_j^{(i)}}$ as the input to the internal classifiers and decide to exit early based on either the entropy of class probabilities or the largest class probability \cite{branchynet, shallowdeep}. We propose feeding the distances to the class means ($d^{k^{(i)}}_j$) as additional inputs to the internal classifiers by simple concatenation, which improves the performance. In addition, during inference, if the ${j^{th}}$ internal classifier decides to move to the next layer, \eename{} is consulted. If \eename{} suggests exiting early, the prediction of the ${j^{th}}$ internal classifier is returned and the input exits early. Hence, an input can move to the next layer if and only if it receives approval from both the internal classifier (which can be based on any existing scheme) and our simple \eename{}.

\subsection{Extension to Unsupervised Learning}
\eename{} can be used for clustering as well. As an example, we consider Deep Embedding Clustering (DEC) \cite{dec} and focus on the task of jointly learning representations and cluster assignments. In DEC, there is only one clustering layer, and it is at the end of the encoder layers. This makes it impractical for low-power clustering, because the architecture is like a tunnel with only one exit at the end. We propose adding multiple clustering layers as early exits in order to decrease the computational cost.

\begin{figure*}[h]
\centerline{\includegraphics[width=0.33\linewidth]{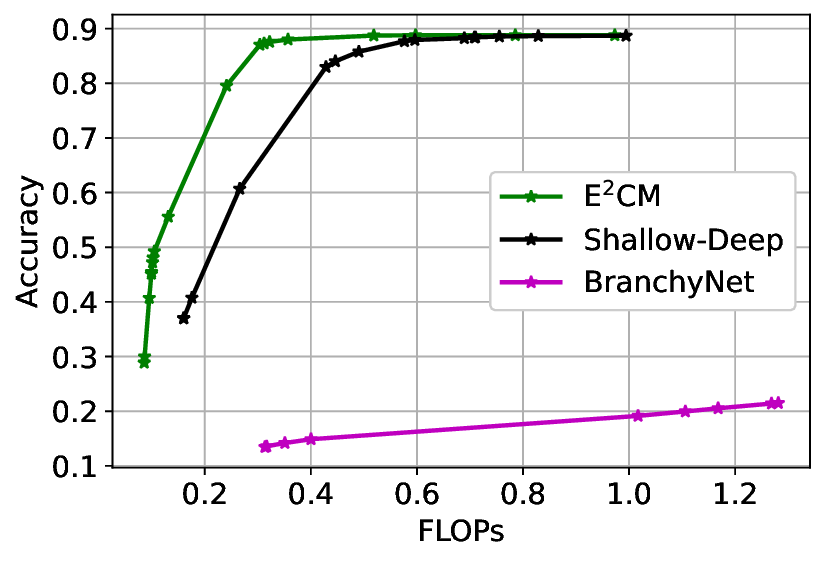}\includegraphics[width=0.33\linewidth]{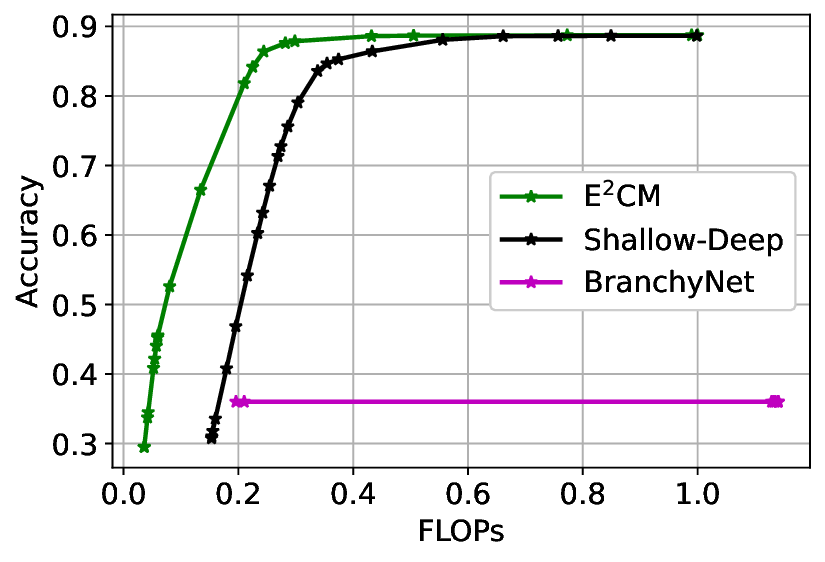}\includegraphics[width=0.33\linewidth]{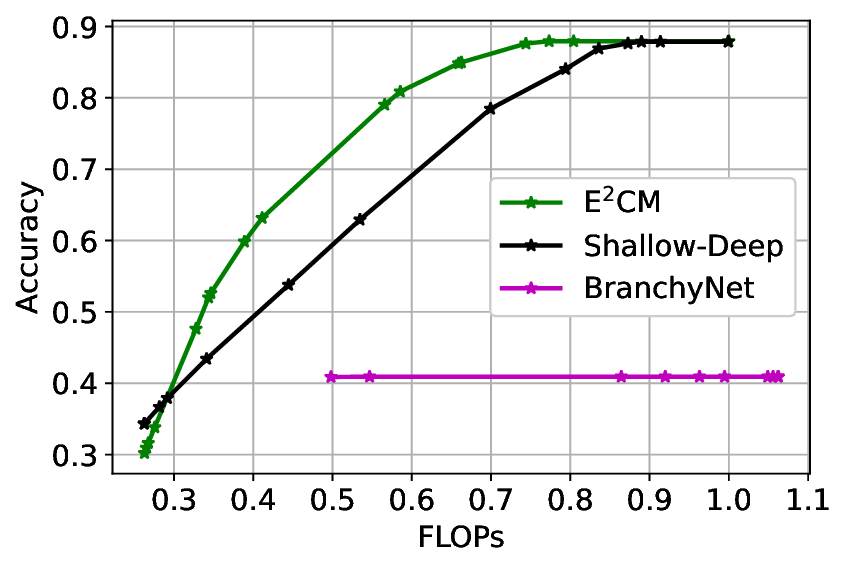}}
\centerline{\includegraphics[width=0.33\linewidth]{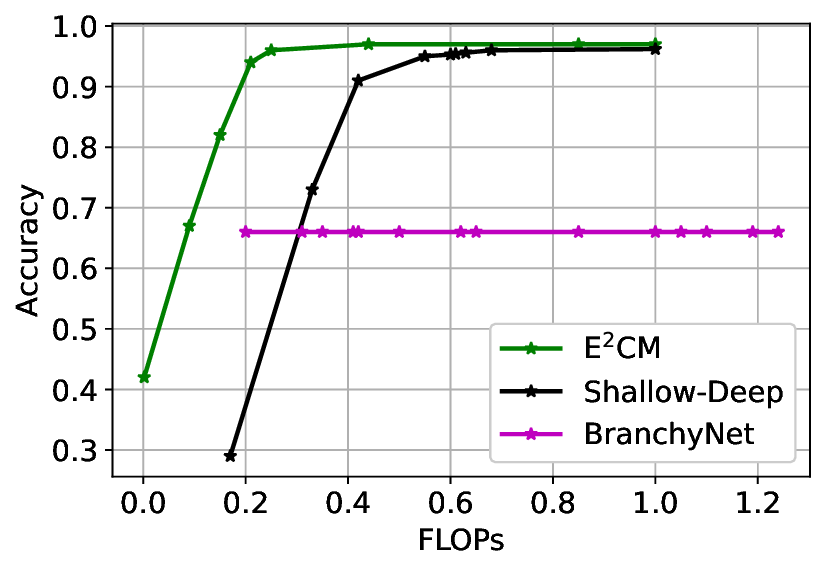}\includegraphics[width=0.33\linewidth]{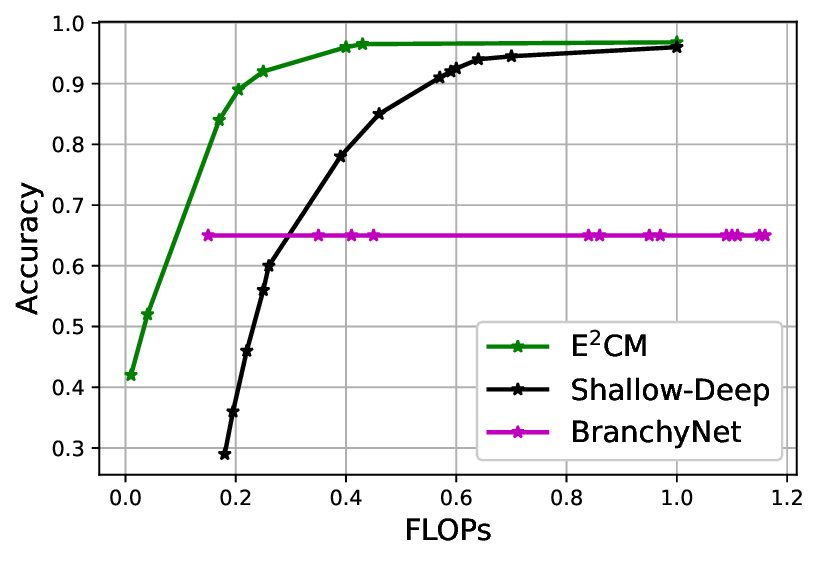}\includegraphics[width=0.33\linewidth]{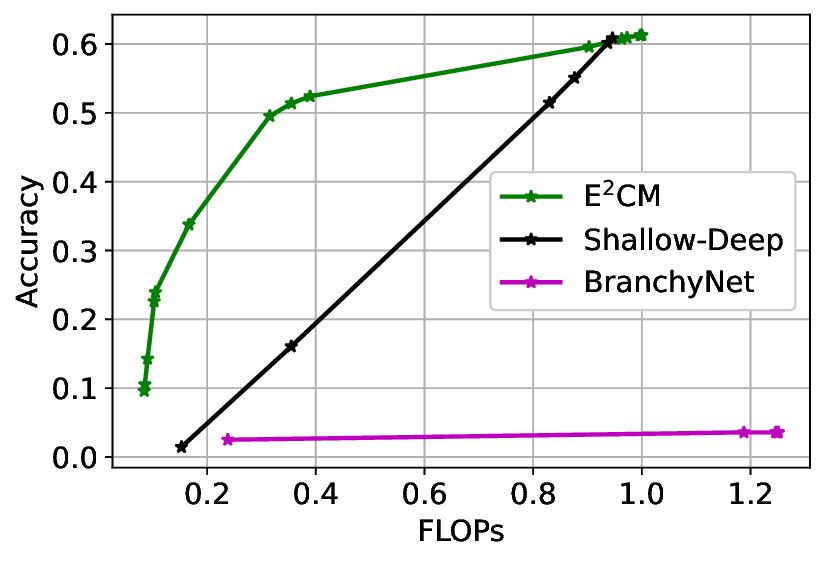}}\vspace{-10pt}
\caption{Comparison of \eename{} with existing methods under fixed training time budget of one epoch for ResNet-152 (left) and WideResNet-101 (middle) on CIFAR-10 (top) and KMNIST (bottom). ResNet-18 for CIFAR-10 is shown at top right, and ResNet-152 for CIFAR-100 is shown at bottom right.}
\label{oneepoch}\vspace{-10pt}
\end{figure*}

\section{Results}
\label{results_label}
We validate the effectiveness of \eename{} on CIFAR-10 \cite{cifar}, CIFAR-100 \cite{cifar}, ImageNet \cite{imagenet}, Tiny ImageNet \cite{tinyimagenet}, KMNIST \cite{kmnist}, Fashion-MNIST \cite{fmnist} and MNIST \cite{mnist} datasets using ResNet-18 \cite{he2016deep}, ResNet-152 \cite{he2016deep}, WideResNet-101 \cite{zagoruyko2016wide}, MobileNetV3Large \cite{mobilenetv3} and EfficientNet-B0 \cite{efficientnet} models.

\textbf{Datasets.} CIFAR-10 and CIFAR-100 datasets consist of 50000 training and 10000 test images, and have 10 and 100 classes respectively with equal amount of samples for each class. ImageNet dataset has 1000 classes and consists of 1.2 million training images and 50000 validation images. Tiny Image-Net dataset consists of 100000 training, 10000 validation and 10000 test images of 200 classes. KMNIST, Fashion-MNIST and MNIST datasets consist of 60000 training and 10000 test images. While MNIST is a dataset of handwritten digits, KMNIST consists of 10 Hiragana characters and Fashion-MNIST consists of clothing images. KMNIST, Fashion-MNIST and MNIST datasets consist of grayscale images unlike other datasets described above.

\textbf{Models and training.} We use ResNet-18, ResNet-152, WideResNet-101, MobileNetV3Large and EfficientNet-B0 models for our supervised learning experiments. For all models, we use the same data augmentation scheme and the hyper-parameter values stated in \cite{veit2018convolutional} for training. For our unsupervised learning experiments, we use the same network architecture and training parameters stated in \cite{dec}.

\textbf{Experiments.} We run experiments in four settings. First, we fix the training time budget and compare \eename{} with the existing early exit methods in terms of network accuracy and floating point operations (FLOPs) performed during inference. The second experiment is similar to the first one: Training time budget is fixed, but now for a fine tuning task. In the third experiment, we remove the training time budget. We allow the full training of the internal classifiers, and we combine \eename{} with internal classifiers and compare it with existing methods. The fourth experiment is for unsupervised learning, and we generate the accuracy-FLOPs curve to evaluate the effectiveness of \eename{}.

\subsection{\eename{} Under a Fixed Training Time Budget}
\label{thresholding_procedure}
In the fixed training time budget setting, we compare \eename{} with existing methods in two ways. First, \eename{} is compared with Shallow-Deep Networks \cite{shallowdeep} and BranchyNet \cite{branchynet}, which are trained for only one epoch since \eename{} requires only a single forward pass. We do not include here the Bolukbasi-Wang-Dekel-Saligrama (BWDS) method \cite{bolukbasi}, because in this method, each decision function requires a separate training. Hence, we cannot train an entire BWDS network with many decision functions using only one epoch.

We use the procedure described in \cite{shallowdeep} and add 6 ICs after the layers which correspond to the 15\%, 30\%, 45\%, 60\%, 75\%, 90\% of the entire network in terms of FLOPs for Shallow-Deep Networks. For BranchyNet, we add 2 branches to the original network as in \cite{branchynet}. The first branch is after the first convolutional layer, and the second branch is after the layer that corresponds to 1/3 of the whole network in terms of FLOPs. For \eename{}, we allow early exiting after every ResNet block for simplicity.

To reach a target number of FLOPs, we first initialize the threshold vector by drawing $M$ numbers uniformly at random from $[0,1]$. The ${j^{th}}$ component of the threshold vector corresponds to ${T_j}$ in Algorithm \ref{alg:classmeans} and is utilized at layer ${l_j}$ of the neural network. Then, the softmax values are obtained for each layer on training set as in \eqref{equation5}. Therefore, only one pass on training set is needed to optimize the thresholds. We update the threshold vector until we reach the target number of FLOPs using binary search at each component and alternating optimization: If the thresholds give larger number of FLOPs than the target FLOP, the thresholds are decreased for the next iteration. Otherwise, they are increased. Binary search is guaranteed to converge as the (average) FLOPS are monotonic with respect to the thresholds.

In this paper, we normalize the FLOPs to the base model. Thus, the base model without any early exits is assumed to cost $1$ FLOP. To obtain the trade-off between accuracy and FLOPs for early exit models synthesized from a base model, the above threshold optimization process is repeated on the training set for target FLOPs ranging from $0.0$ to $1.0$ with $0.001$ granularity. We then compute the convex hull of the resulting $1000$ points in the FLOPs-Accuracy plane. This yields the best performing thresholds on the training set. We use the best thresholds on test set and form the FLOPs-Accuracy curves. Each point on a given curve represents one threshold vector. Points are connected via lines: The performance of any point on any given line is achievable simply via time sharing of the models that correspond to line endpoints.

We show the performance of \eename{} for a fixed training time budget in Fig.~\ref{oneepoch}. The horizontal axis represents the FLOPs normalized the base model, and the vertical axis represents the accuracy. As shown in the figure, \eename{} generally outperforms all existing schemes. Specifically, using \eename{} as the only decision mechanism achieves 50\% better accuracy or 50\% faster inference time for certain cases. This is because \eename{} uses the trained weights of the original vanilla model, which generalizes well to the dataset and can be used for classification. On the other hand, internal classifiers require further training. Therefore, they require more time to be ready for the task of classification. 

\newcommand{\flopvariable}{\phi}

\def\arraystretch{1.5}

\begin{table}[t]
\caption{Comparison of \eename{} with existing methods under a fixed training time budget of six epochs. Symbol $\flopvariable$ represents the FLOP constraint.}
\vspace{-15pt}
\begin{center}
\resizebox{\linewidth}{!}{
\begin{tabular}{|c|c|c|c|c|c|}
\hline
 \large \textbf{Model,}   &\multirow{ 2}{*}{\large\textbf{Method}}    &\multicolumn{4}{c}{\large\bf Accuracy}  \\
\cline{3-6}
 \large \textbf{Dataset}        &     & $\flopvariable\!=\!0.15$ &$\flopvariable\!=\!0.20$ &$\flopvariable\!=\!0.25$ &$\flopvariable\!=\!0.30$\\
\hline
\multirow{3}{*}{ResNet-152,}     &\textbf{\eename{}+SDN}    & \large  \textbf{77\%} &  \large  \textbf{85\%} &  \large  \textbf{87\%} &  \large  \textbf{87\%} \\
\multirow{3}{*}{CIFAR-10}       &SDN                        &   \large  66\% &   \large  75\% &  \large   85\% &    \large 87\% \\
                                &BWDS                       &  \large   77\% &  \large   82\% &    \large 84\% &  \large   86\% \\
                                &BranchyNet                 &  \large    \large 57\% &    \large 57\% &  \large  57\% &   \large 57\% \\
\hline
\multirow{3}{*}{ResNet-152,}     &\textbf{\eename{}+SDN}    & \large  \textbf{84\%} &  \large \textbf{93\%} &  \large \textbf{96\%} &  \large \textbf{96\%} \\
\multirow{3}{*}{CIFAR-10}       &SDN                        &  \large 63\% &  \large 75\% &  \large 86\% &  \large 94\% \\
                                &BWDS                       &  \large 84\% &  \large 93\% &  \large 95\% &  \large 95\% \\
                                &BranchyNet                 &  \large  \large 76\% &  \large 77\% &  \large 78\% &  \large 78\% \\
\hline
\multirow{3}{*}{ResNet-152,}     &\textbf{\eename{}+SDN}    &  \large \textbf{34\%} &  \large \textbf{46\%} &  \large \textbf{52\%} &  \large \textbf{54\%} \\
\multirow{3}{*}{CIFAR-100}      &SDN                        &  \large 20\% &  \large 24\% &  \large 27\% &  \large 30\% \\
                                &BWDS                       &  \large 33\% &  \large 40\% &  \large 42\% &  \large 44\% \\
                                &BranchyNet                 &  \large 24\% &  \large 25\% &  \large 26\% &  \large 26\% \\
\hline
\multirow{3}{*}{WideResNet-101,} &\textbf{\eename{}+SDN}    &  \large 81\% &  \large \textbf{87\%} & \large  \textbf{88\%} &  \large \textbf{88\%} \\
\multirow{3}{*}{CIFAR-10}       &SDN                        &  \large 70\% & \large  79\% &  \large 86\% &  \large 88\% \\
                                &BWDS                       &  \large \textbf{82\%} &  \large 85\% &  \large 87\% &  \large 88\% \\
                                &BranchyNet                 &  \large 56\% &  \large 56\% &  \large 57\% &  \large 57\% \\
\hline
\multirow{3}{*}{WideResNet-101,} &\textbf{\eename{}+SDN}    &  \large 85\% &  \large \textbf{92\%} & \large  \textbf{96\%} &  \large \textbf{96\%} \\
\multirow{3}{*}{KMNIST}         &SDN                        &  \large 63\% &  \large 74\% &  \large 96\% &  \large 96\% \\
                                &BWDS                       &  \large \textbf{88\%} &  \large 90\% &  \large 92\% &  \large 94\% \\
                                &BranchyNet                 &  \large 80\% & \large  82\% &  \large 83\% &  \large 84\% \\
\hline
\end{tabular}}
\end{center}
\label{table_six_epochs}
\vspace{-15pt}
\end{table}

Secondly, we still consider a fixed training time budget, but this time we allow the separate training of decision functions in the BWDS method. As suggested by the authors, there are 6 early exit points, hence 6 decision functions. We train each decision function for 1 epoch, resulting in 6 separate epochs. To make a fair comparison, we train Shallow-Deep Networks and BranchyNet for 6 epochs as well. We combine \eename{} with Shallow-Deep Networks. By doing so, \eename{} rectifies the decisions made by Shallow-Deep Networks. At every exit point of Shallow-Deep Networks, if the decision is made in favor of exiting early at the exit point, \eename{} is consulted. If \eename{} disagrees with Shallow-Deep Networks, early exit does not happen at that exit point.

From Table~\ref{table_six_epochs}, it can be seen that combining \eename{} with Shallow-Deep Networks (SDN) performs better than other methods and can increase the performance by $25\%$ when the number of classes is large. Table~\ref{table_six_epochs} shows accuracies at $\phi\in\{0.15,0.20,0.25,0.30\}$ FLOPs only, because after 0.30 FLOPs, the accuracy stays the same, which indicates that we do not need the tunnel-like design of traditional networks. %This result also suggests that training internal classifiers is not an absolute necessity.

\textbf{Overheads.} 
There are two main sources of overhead for our \eename{} scheme that should be considered. One is the extra computation overhead to calculate the distances to class means. In this context, we would like to note that the results in Fig. \ref{oneepoch} already include the computation overhead as the FLOPs are normalized to the base model. The superior performance of \eename{} suggest that the computation overhead is very low. In fact, for the WideResNet-101 and ResNet-152 models and CIFAR-10 dataset, our experiments have shown that the computational complexity of an early exit block is only around $0.007$ FLOPs. For CIFAR-100 dataset, the overhead is about $0.057$ FLOPs. An issue in the case of a very large number of classes is that the computational complexity of \eename{}  scales linearly with the number of classes. This is also evident with the roughly 10-fold scaling (from $0.007$ to $0.057$) of the computational complexity as soon as one considers CIFAR-100 instead of CIFAR-10. One solution that we will explore in the next set of experiments is to utilize max pooling to reduce the class means' dimensions. We will show that this technique can effectively reduce the computation overhead while still preserving the advantages of \eename{} over existing schemes. 

% It is worth noting that \eename{} results in a maximum of 9\% computational overhead per exit location in terms of FLOPs for WideResNet-101 and ResNet-152 on CIFAR-10. Therefore, the amount of extra compute is not a lot. 

\begin{figure}
\centerline{\includegraphics[width=0.9\linewidth]{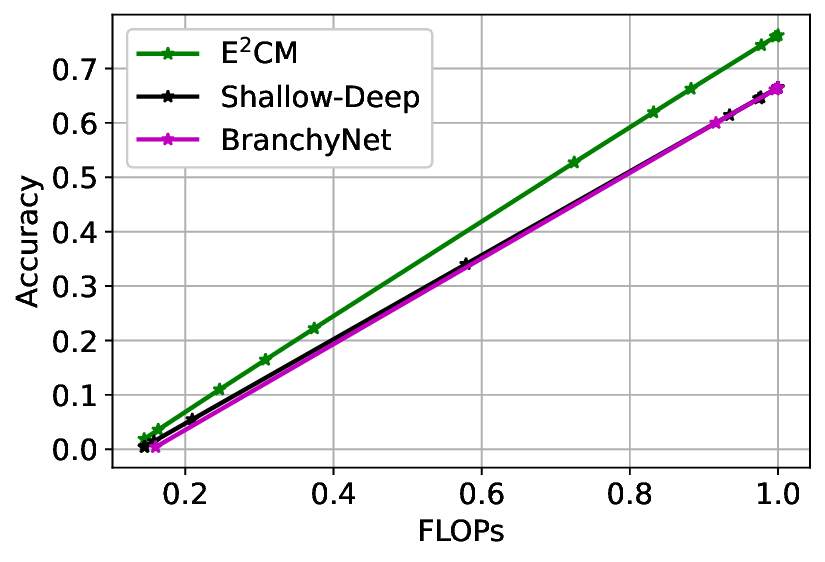}}
\vspace{-5pt}
\centerline{\includegraphics[width=0.9\linewidth]{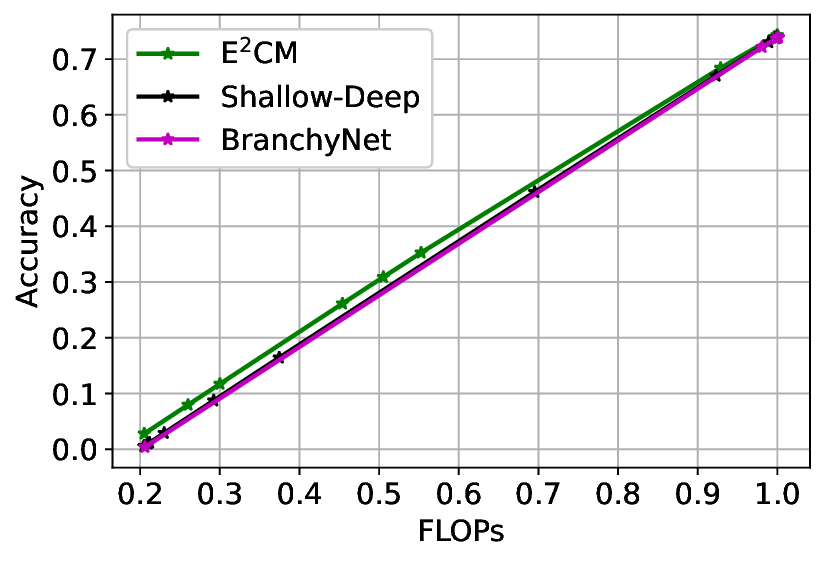}}
\vspace{-5pt}
\centerline{\includegraphics[width=0.9\linewidth]{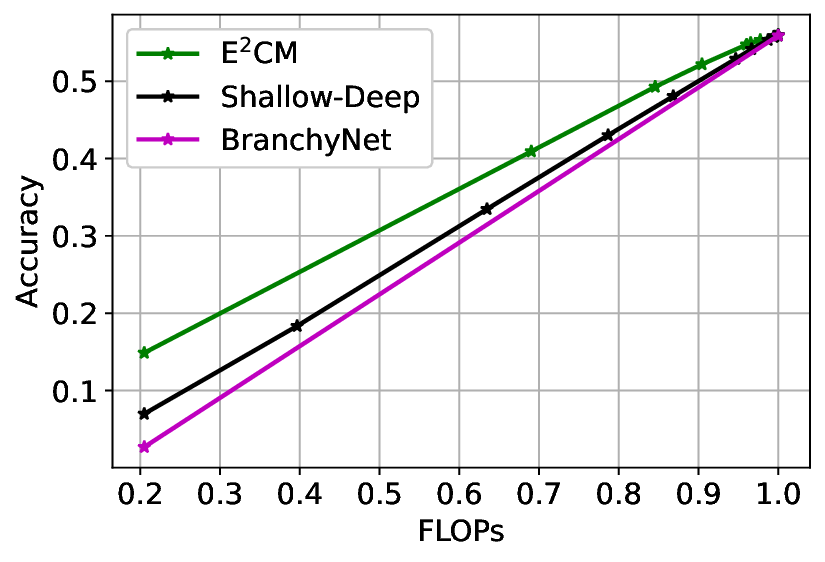}}
\vspace{-10pt}
\caption{Comparison of \eename{} with existing methods under fixed training time budget of one epoch for the fine tuning task on ImageNet (top, middle) and Tiny ImageNet (bottom) datasets using EfficientNet-B0 (top), MobileNetV3Large (middle, bottom) models.}
\vspace{-10pt}
\label{finetuning}
\end{figure}

We have also evaluated the memory overhead of our \eename{} scheme, which is important for edge computing devices with limited resources. Note that, as compared to the base model, \eename{} requires extra memory  for storing the class means vectors. For ResNet-152, WideResNet-101 and ResNet-18 on CIFAR-10 models, we have found that our \eename{} needs $0.417$ MB, $0.43$ MB and $0.242$ MB of space to store the class means for an exit location respectively. These numbers are averages over different layers of the network. Considering there are respectively $50$, $35$ and $8$ ResNet blocks for these models, \eename{} results in a memory overhead of $20.85$ MB, $15.05$ MB and $1.94$ MB respectively. The models themselves respectively require $222$ MB, $476$ MB and $42.17$ MB, so the fractional memory overhead is at most $9.4\%$. Similar to the computation overhead, the memory overhead of \eename{} increases linearly with the number of classes. Thus, for ResNet-152 on CIFAR-100, one will need $0.417 \times 10 = 4.17$ MB to store the class means for a layer on average, increasing the overhead to $9.4\% \times 10 = 94\%$. We show in the following that the aforementioned max pooling method can also alleviate memory requirements. 

%One solution that we will explore in the next set of experiments is to utilize max pooling to reduce the class means' dimensions. This technique can effectively reduce the memory overhead while still preserving its advantages over existing schemes. 

\subsection{\eename{} Under a Fixed Training Time Budget: Fine Tuning} 
In this experiment, we consider a scenario where a model has to be trained on a low-power edge device. Also, the local training dataset is different from the dataset that the base model was trained on. A key motivation for this widely-used transfer learning setup is user privacy.
We compare the early exit methods using EfficientNet-B0 and MobileNetV3Large as base models on ImageNet and Tiny ImageNet datasets. We randomly sample 50 and 10 images per class for ImageNet and Tiny ImageNet and use the resulting subsets for fine tuning. For all methods, only one early exit location is used that corresponds to roughly 20\% of the entire network in terms of FLOPs. To overcome the memory overhead of \eename{}, the output dimensions of the exit layer is downsampled from 14x14x80 to 7x7x10 via max pooling. We follow the same threshold selection procedure as described in \ref{thresholding_procedure}.

As seen from Fig.~\ref{finetuning}, \eename{} outperforms all competing methods under a fixed training time budget for the fine tuning task. This supports the hypothesis that \eename{} generalizes better to the dataset compared to other early exit techniques under a training time budget. This result is important because the networks on edge devices may have to train their own version of the base model  for data/model privacy reasons. Moreover, \eename{} requires the fewest FLOPs for the same accuracy, which translates to reduced battery usage. Moreover, the cost of communication to the cloud may be very high.

% on a local dataset to minimize battery usage and

Thanks to the max pooling technique, the overall computation overhead of \eename{} is only around $0.001$ FLOPs on the ImageNet dataset for both EfficientNet-B0 and MobileNetV3Large. In terms of memory overhead, if MobileNetV3Large is used, \eename{} needs $1.87$ MB and $0.87$ MB to store the class means for ImageNet and Tiny ImageNet datasets respectively. For EfficientNet-B0 on ImageNet, the overhead is $1.87$ MB. Considering MobileNetV3Large requires $21.5$ MB and $17.2$ MB of memory for ImageNet and Tiny ImageNet datasets, respectively, and EfficientNet-B0 needs $16.9$ MB for ImageNet, the memory overhead is at most $11\%$. These results show that \eename{} does not have a large memory or computation footprint even for datasets with a large number of classes. We believe that the memory and computation costs can be further reduced by complementary methods such as quantization and pruning \cite{quantized, compression}. 

\subsection{\eename{} with Unlimited Training}
In this experiment, we remove the training time budget. We combine \eename{} with Shallow-Deep Networks and compare this combination against Shallow-Deep Networks, the BWDS method, and BranchyNet.

\def\arraystretch{1.5}

\begin{table}[]
\caption{Comparison of early exit methods}
\vspace{-15pt}
\begin{center}
\resizebox{\linewidth}{!}{

\begin{tabular}{|c|c|c|c|c|c|}
\hline
\large \textbf{Model,}         &\multirow{2}{*}{ \large \textbf{Method}} &\multicolumn{4}{c}{\large \bf Accuracy} \\
\cline{3-6}
\large \textbf{Dataset} &  &$\flopvariable\!=\!0.15$  &$\flopvariable\!=\!0.20$  &$\flopvariable\!=\!0.25$  &$\flopvariable\!=\!0.30$\\
\hline
\multirow{3}{*}{ResNet-152,}     &\textbf{\eename{}+SDN}    & \large \textbf{82.4\%} & \large \textbf{86.4\%} & \large \textbf{88.5\%} & \large \textbf{88.8\%} \\
\multirow{3}{*}{CIFAR-10}       &SDN                        & \large 80\% & \large 86\% & \large 88.4\% & \large 88.8\% \\
                                &BWDS                       & \large 82\% & \large 84.1\% & \large 86\% & \large 87.7\% \\
                                &BranchyNet                 & \large 80\% & \large 80\% & \large 80\% & \large 80\% \\
\hline
\multirow{3}{*}{ResNet-152,}     &\textbf{\eename{}+SDN}    & \large 94.1\% & \large \textbf{96.2\%} & \large \textbf{96.8\%} & \large \textbf{96.8\%} \\
\multirow{3}{*}{KMNIST}         &SDN                        & \large 83.9\% & \large 94\% & \large 96.4\% & \large 96.6\% \\
                                &BWDS                       & \large \textbf{94.5\%} & \large 96\% & \large 96.5\% & \large 96.6\% \\
                                &BranchyNet                 & \large 86\% & \large 87.5\% & \large 88.2\% & \large 89.5\% \\
\hline
\multirow{3}{*}{WideResNet-101,} &\textbf{\eename{}+SDN}    & \large 84\% & \large \textbf{88\%} & \large \textbf{88.6\%} & \large \textbf{88.8\%} \\
\multirow{3}{*}{CIFAR-10}       &SDN                        & \large 82\% & \large 88\% & \large 88.5\% & \large 88.8\% \\
                                &BWDS                       & \large \textbf{85.8\%} & \large 86.5\% & \large 87.8\% & \large 88.2\% \\
                                &BranchyNet                 & \large 81.9\% & \large 82.5\% & \large 84.9\% & \large 85\% \\
\hline
\multirow{3}{*}{WideResNet-101,} &\textbf{\eename{}+SDN}    & \large \textbf{94.1\%} & \large \textbf{96.5\%} & \large \textbf{97.1\%} & \large \textbf{97.2\%} \\
\multirow{3}{*}{KMNIST}         &SDN                        & \large 84.1\% & \large 94\% & \large 96\% & \large 97\% \\
                                &BWDS                       & \large 93.8\% & \large 94.2\% & \large 95\% & \large 96\% \\
                                &BranchyNet                 & \large 89\% & \large 90.1\% & \large 90.2\% & \large 90.2\% \\
\hline
\end{tabular}}
\end{center}
\label{table_best_epochs}
\vspace{-15pt}
\end{table}

We train the internal classifiers of our merger of \eename{} and Shallow-Deep for 100 epochs. The corresponding high computational complexity may not be desirable for low-power devices. We follow the same threshold selection procedure described above. During inference, the decision of early exit is made according to both the \eename{} and the ICs.

We train BranchyResNet-152 and BranchyWideResNet-101 for 300 and 250 epochs respectively. We use stochastic gradient descent with batch size of 256. The loss of the branches are added up to the loss of the final layer and the weighted average is taken as prescribed in \cite{branchynet}. We have trained multiple combination of weights, and found that $[1/6, 1/4, 1]$ achieves the best performance. For thresholds, we follow the same procedure, but this time the range of values is $[0,\log K]$ where $K=10$, because we consider the entropy of the predictions to make a decision \cite{branchynet}.

Since the task of training decision functions in the BWDS method is a binary classification task (i.e., early exit or not) they converge rather quickly. We separately train the classifiers that come after the decision functions as well. We use pooling and single linear layer for these, as suggested by the authors.

As shown in Table \ref{table_best_epochs}, combining \eename{} with internal classifiers achieves a better trade-off between the computational cost and network accuracy. For low computational budget, \eename{} improves the accuracy by up to 6\%. We also observe that BranchyNet suffers from training the network with the branches jointly. These branches hurt the overall performance. Moreover, convolutional layers in the branches add a significant computational cost without a considerable gain in accuracy. Also, although early exit mechanisms with decision functions like the BWDS method provide decent performance, experimental results show that threshold based early exit mechanisms perform better. In other words, making a decision about early exit and then performing classification fares worse than the threshold-based strategies where classification and exiting decisions are melted into the same pot.

According to Table \ref{table_best_epochs}, we can conclude that the consulting mechanism between the \eename{} and the internal classifiers is generally beneficial. The common decision that is reached by the two classifiers can rectify possible misclassifications and avoid unnecessary computation. This can be seen as an example of ensembles, in which multiple classifiers are used to make a decision. Interestingly, our ensemble reduces the total computational cost unlike ordinary ensemble methods.

\begin{figure} 
\centerline{\includegraphics[width=0.9\linewidth]{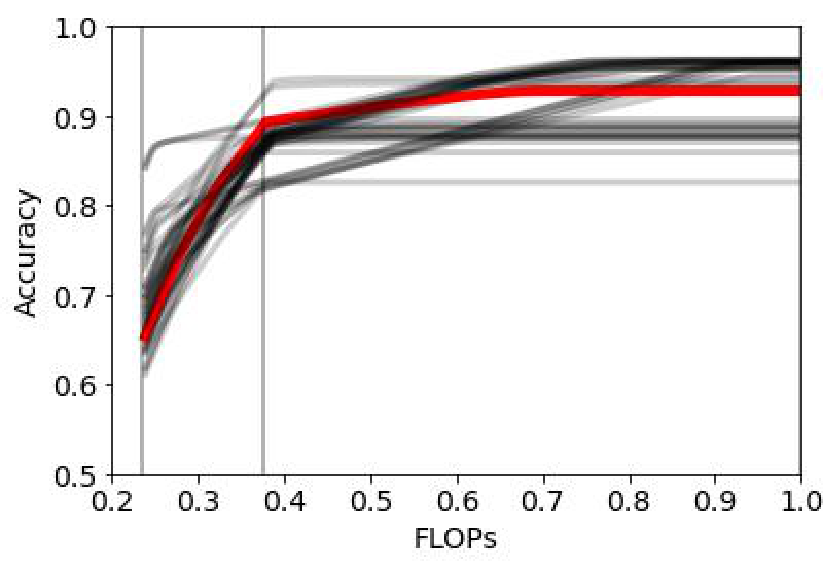}}\vspace{-5pt}
\centerline{\includegraphics[width=0.9\linewidth]{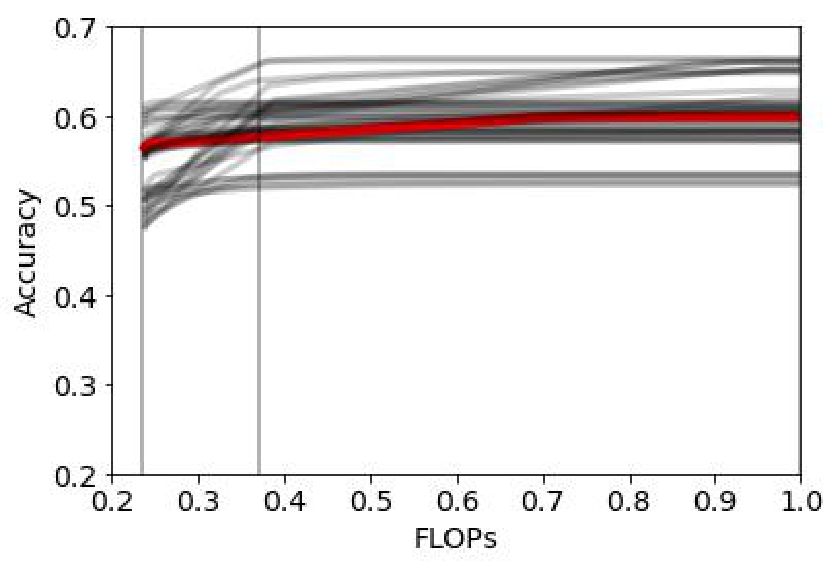}\vspace{-10pt}}
\caption{Accuracy-FLOPs curve on MNIST (top) and Fashion-MNIST (bottom) for unsupervised learning using \eename{}. Vertical lines indicate the individual FLOPs of DECs. Each black curve is for one experiment, and the red curve is the average of all experiments.}\vspace{-10pt}
\label{unsupervised}
\end{figure}

\subsection{\eename{} for Unsupervised Learning}
We follow  the same experimental setup as in \cite{dec} for MNIST and Fashion-MNIST datasets. Namely, we use an encoder with 4 layers, which have 500, 500, 2000, 10 neurons, with a clustering layer (CL) at the end. Let this encoder and clustering layer be $DEC_{large}$. After pretraining $DEC_{large}$, we train the clustering layer as in \cite{dec}. Then, we create another encoder with the 500-500-10-CL architecture, where the weights of first two layers are copied from $DEC_{large}$ and frozen. We name this encoder $DEC_{middle}$ and follow the same procedure as in $DEC_{large}$. Finally, we repeat the same procedure for $DEC_{small}$, which has the 500-10-CL architecture, where the weight of the first layer is frozen and copied from $DEC_{large}$.

After the training is complete, we take the intermediate outputs, i.e., the outputs of the layers with 10 neurons. Then, using the cluster centers from each clustering layer, we follow Algorithm~\ref{alg:classmeans}. To measure the accuracy, we use the same technique described in \cite{dec}.

During the experiments, we noticed that the process of pretraining affected the final result significantly. We have thus run multiple experiments, and the performance corresponding to each experiment is illustrated as one gray curve in Fig.~\ref{unsupervised}. The average of individual experiments is shown as the solid red curve. As can be observed in Fig.~\ref{unsupervised}, by adding early exits to the architecture, it is possible to save $60\%$ of the computation while losing only $6\%$ in unsupervised clustering accuracy on MNIST dataset. On Fashion-MNIST, the accuracy loss is $1\%$. Also, thresholds make it possible to adjust the model according to various computational needs.

\section{Conclusion}
We propose a novel early exit mechanism based on the class means. Unlike existing early exit mechanisms, our method does not modify the base model and does not require gradient-based training, which makes it useful for network training on low-power devices. Under fixed training time budget, our method outperforms existing early exit schemes. In addition, combining our method with existing early exit techniques achieve better trade-off between the computational cost and the network accuracy. Moreover, we show that our method is not only useful in supervised learning tasks, but also in unsupervised learning tasks.

% \label{ethics_statement}
% \subsubsection*{Ethics Statement}
% %\paragraph{Ethics Statement}
% Since our technique reduces the inference time with minimal compromise in performance, it can lead to more severe impacts when used in applications in which time has a crucial importance. Stock market prediction and adversarial attacks that are used in the military industry depend heavily on split second decisions, therefore using \eename{} in such applications may cause negative impacts. Also, as in \cite{shallowdeep}, new adversarial attacks can be developed to manipulate the computational load of deep learning systems which may hurt performance more than before.

% \label{reproducibility}
% \subsubsection*{Reproducibility Statement}
% %\paragraph{Reproducibility Statement}
% We have uploaded the source code for our method as a supplementary material. For better reproducibility, we have utilized common datasets in our experiments, such as CIFAR-10 \cite{cifar}, KMNIST \cite{kmnist}, Fashion-MNIST \cite{fmnist} and MNIST \cite{mnist}. We have also used common model architectures such as ResNet-152 \cite{he2016deep} and WideResNet-101 \cite{zagoruyko2016wide}.

\section*{Acknowledgment}
This work was supported in part by Army Research Lab (ARL) under Grant W911NF-21-2-0272, National Science Foundation (NSF) under Grant CNS-2148182, and by an award from the University of Illinois at Chicago Discovery Partners Institute Seed Funding Program.

\bibliography{ijcnn_conference}
\bibliographystyle{IEEEtran}

\end{document}